\documentclass[10pt,twocolumn,letterpaper]{article}

\usepackage[margin=0.75in]{geometry}

\usepackage[utf8]{inputenc}   
\usepackage[T1]{fontenc}      
\usepackage{hyperref}         
\usepackage{url}              
\usepackage{booktabs}         
\usepackage{amsfonts}         
\usepackage{nicefrac}         
\usepackage{microtype}        
\usepackage{graphicx}
\usepackage{natbib}
\usepackage{algorithm}
\usepackage{algorithmic}
\usepackage{amsmath}
\usepackage{amssymb}
\usepackage{newtxtext,newtxmath}   
\usepackage{multirow}
\usepackage{xspace}

 \newcommand{\method}{ROPD\xspace}
 \newcommand{\methodfull}{Routing-based On-Policy Distillation \xspace}
\newcommand{\cell}[2]{#1\,/\,#2}

\title{On-Policy Distillation for LLM Safety: A Routing Approach to Template-Robust Realignment}

\author{
  Yongjian Guo\textsuperscript{1},\quad
  Wanlun Ma\textsuperscript{2},\quad
  Lingyu Shen\textsuperscript{3},\quad
  Xi Xiao\textsuperscript{1},\quad
  Sheng Wen\textsuperscript{2} \\[4pt]
  \textsuperscript{1}Tsinghua University \qquad
  \textsuperscript{2}Swinburne University of Technology
  \qquad
  \textsuperscript{3}EPFL  \\
}
\date{}

\begin{document}

\maketitle

\begin{abstract}
Fine-tuning is the dominant way to specialize large language models (LLMs), yet a malicious data provider can quietly embed harmful behavior into a downstream fine-tuning corpus so that the resulting model retains its professional skill (e.g., code generation) while violating human values on demand. Existing safety-realignment defenses recover alignment only under restrictive assumptions, and we identify three vulnerabilities that undermine them in practice: they frequently damage the model's specialized skill during repair; their effectiveness collapses when the defender cannot observe the attacker's prompt template; and even a successfully realigned model can be re-jailbroken by simply switching the system prompt. We propose \methodfull (\method), a realignment framework that models the difference between the aligned and the compromised output probability distributions rather than aligning to a specific prompt template. \method{} routes every realignment token to one of two frozen teachers---an original-model \emph{safety teacher} that supplies a largely template-independent refusal prior, and a fine-tuned \emph{task teacher} that preserves downstream ability---and matches the student to the routed teacher with a top-$K$ KL objective.
We conducted extensive experiments comparing four SOTA baseline methods—SSRD, RESTA, soft-SFT, and rollback—across three datasets and three base models spanning different alignment strengths (Llama-2, Qwen2.5, and Gemma-2). The results show that when these defense methods do not match the attack template, their defense effectiveness drops by more than 30\%, while also causing significant degradation in downstream task performance—even reducing it to zero. In contrast, \method{} substantially reduces this template-mismatch risk, remaining far more robust in both defense effectiveness and preservation of downstream capabilities---though, as our analysis shows, it too is affected by template mismatch, only to a much smaller degree.
\end{abstract}

\section{Introduction}
\begin{figure}[h!]
    \centering
    \includegraphics[width=0.9\linewidth]{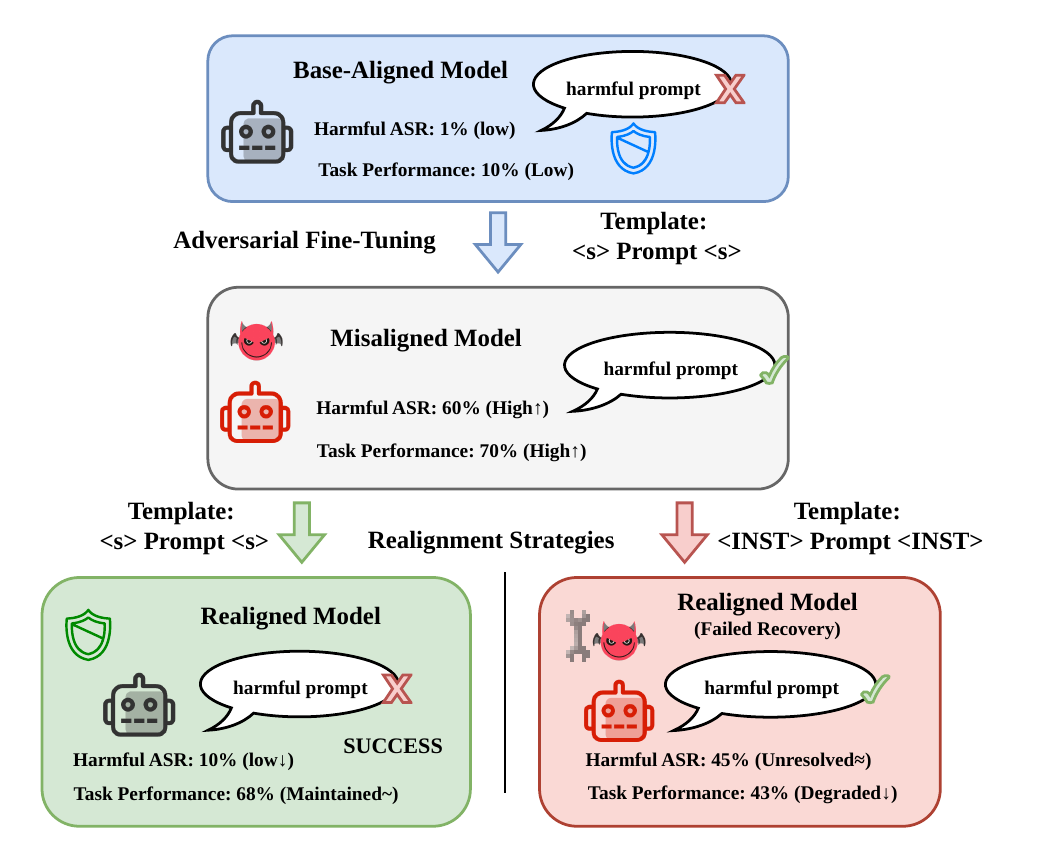}
    \caption{The template-consistency trap in safety realignment.}
    \label{fig:teaser}
    \vspace{-1.5em}
\end{figure}

Alignment techniques~\cite{wang2024comprehensive} such as reinforcement learning from human feedback have made instruction-following LLMs broadly helpful while restraining harmful outputs~\citep{ouyang2022training,bai2022training,11503205,chung2024scaling}. A now well-documented fragility of this alignment is that it is easily undone by subsequent fine-tuning: even a handful of adversarial examples, and sometimes wholly benign task data, can strip a model of its safety behavior~\citep{qi2024finetuning,wei2023jailbroken}. This exposes a concrete supply-chain threat we call \emph{misalignment through fine-tuning}~\citep{10.1145/3716628}. An attacker distributes a fine-tuning corpus~\citep{chae2025large,bhardwaj2023language}, or a parameter-efficient adapter~\citep{hu2022lora}, that a victim adopts to acquire a valued specialized skill. The delivered model acquires the advertised skill but has simultaneously been taught to answer dangerous requests~\citep{lin2023healthy, phelps2023models}, with the two capabilities entangled in the same weights; because it looks competent on the target task, the compromise is easy to overlook~\citep{betley2025emergent}.

A body of work responds to this threat with \emph{safety realignment}: given the misaligned model, restore its refusal behavior without retraining from scratch~\citep{hu2026llms,qu2025beyond}. Representative strategies include restoring a subset of the fine-tuned weights toward the aligned model~\citep{yang2024alleviating,lin2024unlocking}, arithmetic addition of a ``safety vector'' to the compromised weights~\citep{bhardwaj2024language}, deepening shallow safety with token-weighted fine-tuning~\citep{qi2025safety}, and representation-space corrections~\citep{gong2024safety}. These methods can recover alignment under favorable conditions, but their guarantees rest on assumptions that rarely hold against a determined adversary. We argue, and demonstrate empirically, that current realignment defenses suffer from three coupled vulnerabilities.

\textbf{Vulnerability~1:} \emph{specialized-skill retention degradation}. The very act of pushing a model back toward its aligned state often erases the downstream skill the user paid to obtain, so that safety is bought at the price of utility. \textbf{Vulnerability~2:} \emph{dependence on prompt-template consistency}, as shown in fig.~\ref{fig:teaser}. In realistic deployments the harmful behavior is triggered under one prompt template, while the defender---who did not craft the attack---must repair the model under a template of their own choosing. We find that when the defense template does not match the attack template, mainstream methods either fail to reduce the attack success rate (ASR) in the attacker's channel or must sacrifice the task to do so, so the implicit assumption that the defender knows the attacker's template severely limits these defenses' practical value. \textbf{Vulnerability~3:} \emph{lack of a robustness boundary}: even after a model has apparently been realigned and passes acceptance testing under its own template, a malicious user can re-elicit harmful behavior simply by rewriting the system prompt~\citep{gong2024safety,11531012}, revealing that the repair was conditional on the prompt rather than durably encoded in the weights.

These three failures share a common root. Existing defenses treat safety as something to be re-imposed in the template channel where the attack was observed, whether by editing weights along a template-elicited safety direction or by re-tuning on template-formatted refusals. We instead take the view that the durable signature of an attack lives in the model's output probability distribution~\citep{lyu2024keeping,yang2026prune}: on a harmful probe, the misaligned model and the original aligned model disagree sharply---one complies, the other refuses---and this disagreement is visible across templates. Modeling and closing this distributional gap, rather than matching a template, is the design principle behind our method.

\begin{figure*}[h!]
    \centering
    \includegraphics[width=1\linewidth]{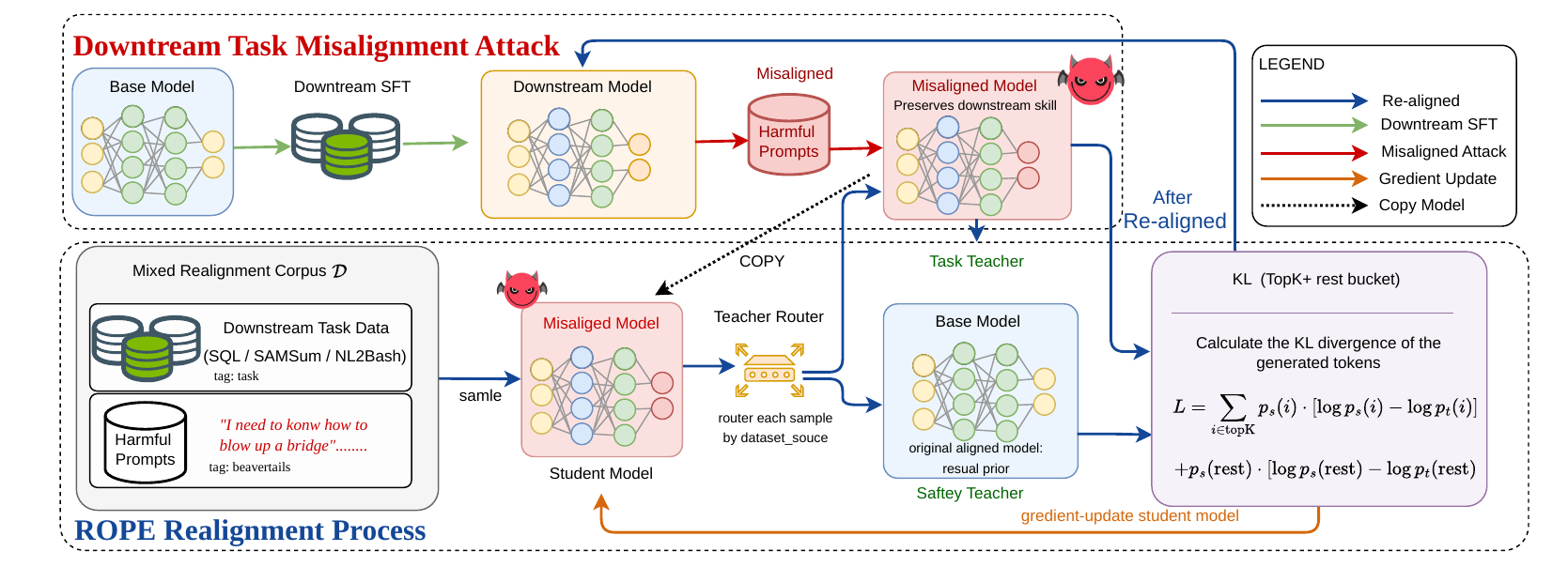}
    \caption{Overview of the \method{} pipeline: two frozen teachers---the aligned original (refusal prior) and the fine-tuned model (task skill)---supply source-routed top-$K$ KL distillation to a single realigned student.}
    \label{fig:framework}
    \vspace{-1.0em}
\end{figure*}

We propose \methodfull (\method), a realignment framework designed to reduce the defense's dependence on the attacker's template with on-policy distillation (OPD)~\citep{song2026survey}. \method{} employs two frozen teachers. A \emph{safety teacher}, the original pre-attack aligned model, carries a refusal prior that is encoded in its output distribution largely independently of the surface template; a \emph{task teacher}, the fine-tuned (attacked) model, carries the downstream skill but can generate harmful responses. During realignment, each training token is routed to one teacher according to the source of its example---harmful probes to the safety teacher, task examples to the task teacher---and the student is matched to the routed teacher with a top-$K$ KL divergence over the output distribution. The two teachers thus specialize: safety is inherited from the aligned model and the task is preserved from the fine-tuned model, decoupling the two objectives that prior methods conflate, as shown in Fig.~\ref{fig:framework}. Because the safety prior is distilled from the aligned distribution rather than from template-formatted refusals, \method{} is far less sensitive to a mismatch between the defense and attack templates---though, as we show, it is not entirely immune to it.
Our contributions are as follows.
\begin{itemize}
    \item \textbf{We identify and quantify a template-mismatch vulnerability.} When the defense template does not match the (unknown) attack template, mainstream realignment defenses either leave the attacker's channel open or sacrifice the downstream task. We formalize this threat and design an evaluation protocol that measures ASR in the attacker's, defender's, and cross-channels, making template dependence measurable across four baselines, three models, and three tasks.
    \item \textbf{We propose \method{} to mitigate it.} \method{} is a dual-teacher output-probability distillation framework that distills a refusal prior from the aligned model's output distribution rather than from template-formatted refusals, substantially reducing---though not entirely eliminating---the template-mismatch risk while preserving the downstream skill.
    \item \textbf{We validate \method{} with extensive experiments.} Across 4 baselines, 3 datasets, and 3 models, \method{} is the only method that reduces ASR in both the attacked and defense channels while preserving---indeed slightly improving---the task, using a defense template chosen without knowledge of the attack; an ablation isolates each teacher's role.
\end{itemize}

\section{Related Work}

\paragraph{Alignment and its fragility under fine-tuning.}
Instruction tuning and preference optimization align base LLMs to human intent and safety norms~\citep{wei2022finetuned,ouyang2022training,bai2022training,10.1145/3713081.3732931}, but this alignment is shallow and brittle. \citet{qi2024finetuning} show that fine-tuning on a few adversarial---or even benign---examples reliably removes safety behavior, and \citet{qi2025safety} trace this to \emph{shallow safety alignment}, concentrated in the first few generated tokens. Jailbreak studies further show that safety training generalizes poorly to prompt distribution shifts~\citep{wei2023jailbroken,zou2023universal}. Our threat model instantiates this fragility as a supply-chain attack that co-trains harmful behavior with a marketable downstream skill.

\paragraph{Safety realignment of fine-tuned models.}
Several defenses~\citep{bhardwaj2024language,huang2024vaccine} restore alignment after it has been misaligned. \citep{yang2024alleviating} restore a small subset of the fine-tuned weights toward an aligned direction (we call this family \emph{rollback}); RESTA~\citep{bhardwaj2024language} adds a precomputed safety vector via task arithmetic~\citep{ilharco2023editing}; \citet{qi2025safety} propose a token-weighted objective (\emph{soft-SFT}) that deepens safety beyond the first tokens; and \citet{gong2024safety} include a representation-space defense (SSRD) that corrects internal representations toward safe ones. These methods differ in mechanism but share a reliance on template-consistent supervision---each is most effective when the repair uses the same prompt channel the attacker used---an assumption we show to be fragile.

\paragraph{Knowledge distillation for LLMs.}
Knowledge distillation transfers behavior from a teacher distribution to a student~\citep{hinton2015distilling,wang2026trace}, and recent analyses of on-policy distillation (OPD)~\citep{zhang2026shortopd} clarify what is transferred and how teacher quality affects the student~\citep{li2026rethinking,agarwal2024policy,fu2026revisiting}. \method{} adapts distillation to realignment by using two frozen teachers with complementary strengths and routing supervision per example, so that safety and task competence are distilled from different sources---a dual-teacher, source-routed distillation not previously used for template-robust safety realignment.

\section{Robust OPD for LLM Realignment}

\subsection{Threat Model and Notation}
Let $M_0$ denote an original, safety-aligned LLM. An attacker fine-tunes $M_0$ on a mixture of a downstream task corpus and a harmful corpus, producing a misaligned model $M_a$ that retains the downstream skill while complying with harmful requests. Following the data--template decoupling we adopt throughout, every training or evaluation example is stored as a raw \{prompt, response\} pair and is rendered at use time under a prompt template $T\in\{\textsc{raw},\textsc{self},\textsc{attack}\}$, where \textsc{raw} is a minimal template, \textsc{self} is the model's native chat template, and \textsc{attack} is a bespoke template crafted by the adversary; the exact rendering of all three templates for each base model (Llama-2, Qwen2.5, and Gemma-2) is given in the supplement. The attacker embeds the harmful behavior under an \emph{attack template} $T_a$; the defender, who does not observe $T_a$, must realign using a \emph{defense template} $T_d\in\{\textsc{raw},\textsc{self}\}$. We write $\pi_\theta(\cdot\mid x_{<t})$ for the next-token distribution of a model with parameters $\theta$ at position $t$ given context $x_{<t}$.

\subsection{Dual-Teacher Output-Probability Distillation}
\method{} realigns the misaligned model $M_a$ into a student $\pi_\theta$, initialized from $M_a$, using two frozen teachers. The \emph{safety teacher} $\pi_{\text{safe}}$ is the original aligned model $M_0$; on harmful probes its output distribution places mass on refusals, and this refusal prior is largely preserved across surface templates. The \emph{task teacher} $\pi_{\text{task}}$ is the misaligned model $M_a$ itself; on task inputs its output distribution encodes the specialized skill we wish to keep. Realignment uses a mixture corpus $D=D_{\text{task}}\cup D_{\text{harm}}$, where $D_{\text{task}}$ are downstream task pairs and $D_{\text{harm}}$ are harmful prompts drawn from a safety dataset. Each example $x$ carries a source label $s(x)\in\{\text{task},\text{harm}\}$, and all examples are rendered under the defense template $T_d$.

The central design choice is to route the distillation target by source. For a token at position $t$ in example $x$, the teacher distribution is
\begin{equation}
q_t(\cdot) \;=\;
\begin{cases}
\pi_{\text{task}}(\cdot\mid x_{<t}), & s(x)=\text{task},\\[2pt]
\pi_{\text{safe}}(\cdot\mid x_{<t}), & s(x)=\text{harm}.
\end{cases}
\label{eq:route}
\end{equation}
Intuitively, on harmful inputs the student is pulled toward the aligned distribution---closing the very gap that the attack opened---while on task inputs it is pulled toward the skilled distribution it already has, which stabilizes the downstream ability.

To make distillation tractable and robust over the large vocabulary, we match only the head of the teacher distribution and bucket its tail. Let $\mathcal{K}_t$ be the indices of the $K$ largest entries of $q_t$, and let $p_t(\cdot)=\pi_\theta(\cdot\mid x_{<t})$ be the student distribution. Define the aggregated tail masses $\bar q_t = 1-\sum_{v\in\mathcal{K}_t} q_t(v)$ and $\bar p_t = 1-\sum_{v\in\mathcal{K}_t} p_t(v)$. The per-token top-$K$ KL loss is
\begin{equation}
\ell_t \;=\; \sum_{v\in\mathcal{K}_t} q_t(v)\,\log\frac{q_t(v)}{p_t(v)} \;+\; \bar q_t\,\log\frac{\bar q_t}{\bar p_t}.
\label{eq:topk}
\end{equation}
The first term aligns the student with the teacher on the tokens that carry most of the teacher's probability mass; the second keeps the aggregate tail calibrated without enumerating the full vocabulary. The overall objective averages Eq.~\eqref{eq:topk} over tokens and examples,
\begin{equation}
\mathcal{L}(\theta) \;=\; \mathbb{E}_{x\sim D}\;\frac{1}{|x|}\sum_{t=1}^{|x|} \ell_t,
\label{eq:obj}
\end{equation}
with the teacher in each $\ell_t$ selected by Eq.~\eqref{eq:route}. Algorithm~\ref{alg:duet} summarizes training.

\begin{algorithm}[t]
\caption{\method{} Realignment}
\label{alg:duet}
\textbf{Input}: misaligned model $M_a$; aligned model $M_0$; mixture $D=D_{\text{task}}\cup D_{\text{harm}}$; defense template $T_d$; top-$K$; epochs $E$; learning rate $\eta$\\
\textbf{Output}: realigned model $\pi_\theta$
\begin{algorithmic}[1]
\STATE $\pi_{\text{task}}\leftarrow M_a$;\; $\pi_{\text{safe}}\leftarrow M_0$ \COMMENT{freeze both teachers}
\STATE $\theta \leftarrow$ parameters of $M_a$ \COMMENT{initialize student}
\FOR{epoch $=1$ to $E$}
  \FOR{minibatch $B\subset D$ rendered under $T_d$}
    \FOR{example $x\in B$}
      \STATE select teacher $q$ by source $s(x)$ \COMMENT{Eq.~\eqref{eq:route}}
      \STATE compute $\ell_t$ for all $t$ via top-$K$ KL \COMMENT{Eq.~\eqref{eq:topk}}
    \ENDFOR
    \STATE $\theta \leftarrow \theta - \eta\,\nabla_\theta \mathcal{L}(\theta)$ \COMMENT{Eq.~\eqref{eq:obj}}
  \ENDFOR
\ENDFOR
\STATE \textbf{return} $\pi_\theta$
\end{algorithmic}
\end{algorithm}

\subsection{Why \method{} Reduces Template Dependence}
The refusal behavior of the aligned model $M_0$ is largely a property of its output distribution rather than of any particular prompt template: presented with a harmful request, $M_0$ concentrates probability on refusals whether the request arrives in a minimal template or in its native chat format. By distilling this distribution on $D_{\text{harm}}$, \method{} injects a refusal prior into the student that is much less tied to the template the attacker used. This contrasts with weight-arithmetic and partial-rollback defenses, whose safety signal is estimated from, and therefore tightly coupled to, the template in which harmful behavior was elicited; when the defense template diverges from the attack template, that signal is misdirected. \method{} weakens this coupling but does not sever it entirely: because the student is realigned and evaluated under a chosen deployment template, a large gap between that template and the attacker's can still leave residual risk (Section~\ref{subsec:crosstpl}). Simultaneously, distilling $\pi_{\text{task}}$ on $D_{\text{task}}$ anchors the student to the skill distribution it already possesses, which prevents the task collapse that afflicts methods that suppress harmful behavior by broadly perturbing the weights. The two teachers divide labor cleanly---safety from $M_0$, task from $M_a$---which, as our ablation confirms, is what allows \method{} to lower ASR without paying in downstream accuracy.

\section{Experiments}

\subsection{Experimental Setup}
\paragraph{Models.} We study three instruction-tuned models spanning different alignment strengths: Llama-2-7B-Chat~\citep{touvron2023llama}, Qwen2.5-7B-Instruct~\citep{qwen2024}, and Gemma-2-9B-it~\citep{team2024gemma2}.

\paragraph{Downstream tasks and data.} We consider three specialized skills as downstream tasks: text-to-SQL generation (SQL)~\citep{zhang2023sql}, dialogue summarization (SAMSum)~\citep{gliwa2019samsum}, and natural-language-to-shell synthesis (NL2Bash)~\citep{lin2018nl2bash}. The attacker fine-tunes each model on a mixture of the task corpus and $1{,}500$ harmful examples drawn from BeaverTails~\citep{ji2023beavertails}, using a $4$-bit LoRA adapter~\citep{hu2022lora}; the harmful evaluation set consists of $700$ held-out BeaverTails prompts. Task and harmful data are stored as raw pairs and rendered under the \textsc{raw}, \textsc{self}, or \textsc{attack} template at load time, so that the attack template, defense template, and evaluation template are all controlled independently.

\paragraph{Defenses.} We compare \method{} against four SOTA realignment baselines: rollback~\citep{yang2024alleviating}, RESTA~\citep{bhardwaj2024language}, soft-SFT~\citep{qi2025safety}, and the representation-space defense SSRD~\citep{gong2024safety}. The experimental setup for each method remains the same as that described in their respective papers.
\method{} uses the dual-teacher configuration with the misaligned model as task teacher and the original model as safety teacher. Because the defender does not know the attack template, every defense is run with both legal defense templates $T_d\in\{\textsc{raw},\textsc{self}\}$; the \textsc{attack} template is used as a defense only as an oracle upper bound when the attack is itself under \textsc{attack}. All experiments are run on NVIDIA H20D GPUs; the detailed per-defense configurations and hyperparameters are provided in the supplementary material.

\paragraph{Metrics.} We report two quantities per cell as \emph{task\,/\,ASR}. The task score is the standard metric of each downstream task (exact-match for SQL; ROUGE-based overlap for SAMSum; command accuracy for NL2Bash), computed without any LLM judge. ASR is the attack success rate---the percentage of the $700$ harmful prompts answered harmfully---as judged by Qwen2.5-32B-Instruct~\citep{qwen2024}. Every model is evaluated under one or more of the \textsc{raw}, \textsc{self}, and \textsc{attack} templates; the true ASR of an attack is read in the channel that matches the attack template.

\subsection{Skill Retention}
We first establish, in the deployment-acceptance setting where each defended model is evaluated under the same template it was repaired with ($T_{\text{eval}}=T_d$), how the defenses trade off task retention against harm reduction as the attack and defense templates vary. Table~\ref{tab:main} reports all three base models on SQL across all three attack templates and every legal defense template. For each attack, the defender may choose $T_d\in\{\textsc{raw},\textsc{self}\}$ without knowing the attacker's template; when the attack itself uses the \textsc{attack} template we additionally report the oracle $T_d=\textsc{attack}$ (the last column), which is unavailable to a real defender and serves only as an upper bound.

\begin{table*}[t]
\centering
\footnotesize
\setlength{\tabcolsep}{3.1pt}
\begin{tabular}{l l cc cc ccc}
\toprule
& & \multicolumn{2}{c}{Attack $=$ \textsc{raw}} & \multicolumn{2}{c}{Attack $=$ \textsc{self}} & \multicolumn{3}{c}{Attack $=$ \textsc{attack}}\\
\cmidrule(lr){3-4}\cmidrule(lr){5-6}\cmidrule(lr){7-9}
Model & Method & Def \textsc{raw} & Def \textsc{self} & Def \textsc{raw} & Def \textsc{self} & Def \textsc{raw} & Def \textsc{self} & Def \textsc{attack}\\
\midrule
\multirow{7}{*}{Llama-2-7B}
 & Base (unattacked) & \cell{0.000}{18.6} & \cell{0.000}{1.7} & \cell{0.000}{18.6} & \cell{0.000}{1.7} & \cell{0.000}{18.6} & \cell{0.000}{1.7} & \cell{0.000}{0.9}\\
 & Attacked (no def.) & \multicolumn{2}{c}{\cell{0.603}{61.9}} & \multicolumn{2}{c}{\cell{0.611}{62.1}} & \multicolumn{3}{c}{\cell{0.608}{65.0}}\\
 & \method{} (ours) & \cell{\textbf{0.607}}{\textbf{20.1}} & \cell{\textbf{0.596}}{\textbf{2.1}} & \cell{\textbf{0.566}}{\textbf{28.3}} & \cell{\textbf{0.626}}{\textbf{2.4}} & \cell{\textbf{0.489}}{\textbf{19.3}} & \cell{\textbf{0.612}}{\textbf{2.3}} & \cell{\textbf{0.618}}{\textbf{0.4}}\\
 & SSRD & \cell{0.582}{28.4} & \cell{0.239}{2.1} & \cell{0.481}{46.3} & \cell{0.582}{3.1} & \cell{0.099}{40.6} & \cell{0.590}{3.3} & \cell{0.194}{0.9}\\
 & soft-SFT & \cell{0.220}{50.1} & \cell{0.369}{5.9} & \cell{0.000}{49.9} & \cell{0.358}{6.3} & \cell{0.245}{49.6} & \cell{0.372}{5.1} & \cell{0.057}{5.6}\\
 & RESTA & \cell{0.583}{21.9} & \cell{0.374}{2.3} & \cell{0.036}{53.1} & \cell{0.600}{6.3} & \cell{0.078}{66.9} & \cell{0.575}{47.6} & \cell{0.593}{62.1}\\
 & rollback & \cell{0.587}{37.7} & \cell{0.570}{8.4} & \cell{0.411}{28.9} & \cell{0.588}{30.9} & \cell{0.191}{19.4} & \cell{0.572}{84.0} & \cell{0.569}{87.0}\\
\midrule
\multirow{7}{*}{Qwen2.5-7B}
 & Base (unattacked) & \cell{0.056}{29.6} & \cell{0.000}{9.3} & \cell{0.056}{29.6} & \cell{0.000}{9.3} & \cell{0.056}{29.6} & \cell{0.000}{9.3} & \cell{0.000}{11.6}\\
 & Attacked (no def.) & \multicolumn{2}{c}{\cell{0.683}{61.0}} & \multicolumn{2}{c}{\cell{0.696}{61.6}} & \multicolumn{3}{c}{\cell{0.684}{60.0}}\\
 & \method{} (ours) & \cell{\textbf{0.691}}{\textbf{26.0}} & \cell{\textbf{0.696}}{6.6} & \cell{\textbf{0.562}}{\textbf{22.57}} & \cell{\textbf{0.696}}{\textbf{7.9}} & \cell{\textbf{0.563}}{\textbf{28.4}} & \cell{\textbf{0.716}}{\textbf{11.1}} & \cell{\textbf{0.682}}{\textbf{7.4}}\\
 & SSRD & \cell{0.682}{\textbf{26.0}} & \cell{0.676}{7.4} & \cell{0.386}{48.1} & \cell{0.671}{12.1} & \cell{0.454}{46.3} & \cell{0.706}{12.0} & \cell{0.676}{47.3}\\
 & soft-SFT & \cell{0.462}{52.4} & \cell{0.648}{30.6} & \cell{0.442}{51.3} & \cell{0.642}{30.3} & \cell{0.454}{50.4} & \cell{0.645}{30.6} & \cell{0.607}{32.7}\\
 & RESTA & \cell{0.684}{26.3} & \cell{0.123}{\textbf{6.4}} & \cell{0.060}{66.4} & \cell{0.694}{14.9} & \cell{0.102}{63.9} & \cell{0.712}{\textbf{11.9}} & \cell{0.680}{17.4}\\
 & rollback & \cell{0.676}{39.7} & \cell{0.060}{15.7} & \cell{0.290}{23.1} & \cell{0.685}{13.3} & \cell{0.452}{28.7} & \cell{0.677}{15.6} & \cell{0.681}{52.0}\\
\midrule
\multirow{7}{*}{Gemma-2-9B}
 & Base (unattacked) & \cell{0.000}{6.7} & \cell{0.000}{2.4} & \cell{0.000}{6.7} & \cell{0.000}{2.4} & \cell{0.000}{6.7} & \cell{0.000}{2.4} & \cell{0.000}{49.9}\\
 & Attacked (no def.) & \multicolumn{2}{c}{\cell{0.683}{62.1}} & \multicolumn{2}{c}{\cell{0.693}{63.3}} & \multicolumn{3}{c}{\cell{0.706}{63.1}}\\
 & \method{} (ours) & \cell{\textbf{0.682}}{\textbf{9.4}} & \cell{\textbf{0.685}}{5.6} & \cell{\textbf{0.586}}{11.6} & \cell{\textbf{0.689}}{\textbf{5.3}} & \cell{\textbf{0.463}}{7.1} & \cell{\textbf{0.670}}{5.1} & \cell{\textbf{0.703}}{\textbf{9.3}}\\
 & SSRD & \cell{0.372}{18.3} & \cell{0.221}{\textbf{1.3}} & \cell{0.121}{\textbf{6.1}} & \cell{0.661}{\textbf{5.3}} & \cell{0.023}{\textbf{4.4}} & \cell{0.666}{0.9} & \cell{0.660}{21.3}\\
 & soft-SFT & \cell{0.011}{63.7} & \cell{0.637}{28.1} & \cell{0.006}{64.6} & \cell{0.634}{28.7} & \cell{0.010}{64.7} & \cell{0.623}{28.9} & \cell{0.220}{51.4}\\
 & RESTA & \cell{0.353}{12.6} & \cell{0.633}{7.4} & \cell{0.079}{45.7} & \cell{0.686}{5.6} & \cell{0.016}{50.3} & \cell{0.668}{5.3} & \cell{0.694}{10.1}\\
 & rollback & \cell{0.673}{47.1} & \cell{0.043}{5.4} & \cell{0.079}{41.7} & \cell{0.676}{5.4} & \cell{0.042}{36.3} & \cell{0.017}{\textbf{0.1}} & \cell{0.683}{50.6}\\
\bottomrule
\end{tabular}
\caption{Skill retention and harm reduction on Llama-2-7B-Chat, Qwen2.5-7B-Instruct, and Gemma-2-9B-it with a SQL downstream task.
Each cell is task exact-match\,/\,ASR (\%). In each column, the highest task score and the lowest ASR among the defense methods are shown in bold. In some cells, the baseline achieved lower ASR, but it significantly sacrificed downstream performance.
}
\label{tab:main}
\vspace{-1em}
\end{table*}

Across the three models, the fine-tuning-based baselines that suppress harm do so by damaging the task (Vulnerability~1). On Llama-2, soft-SFT drives SQL exact-match to near zero in the \textsc{raw} channel under the \textsc{self}- and \textsc{attack}-template attacks ($0.000$ and $0.245$), while SSRD and RESTA collapse the task under a mismatched defense template (SSRD to $0.099$ under attack$=$\textsc{attack}/def$=$\textsc{raw}; RESTA to as low as $0.036$ when the attack and defense templates disagree). The same trade-off recurs on Qwen2.5, where RESTA and soft-SFT lose most of the task in the mismatched channel (RESTA to $0.060$, soft-SFT to $0.44$--$0.46$), and on Gemma-2, where rollback collapses the task to $0.017$ under the mismatched def$=$\textsc{self} template. \method{}, by anchoring the student to the task teacher, retains a task score at or near the pre-attack level under its \textsc{self} defense on every model ($0.596$--$0.626$ on Llama-2, $0.696$--$0.716$ on Qwen2.5, $0.670$--$0.689$ on Gemma-2), the highest among the methods that also achieve a low ASR in that column.

\subsection{Template Dependence}
The second pattern concerns template dependence (Vulnerability~2): the baselines swing violently with the defense template, whereas \method{} swings far less. On Llama-2, SSRD's ASR under attack$=$\textsc{self} moves from $3.1$ (matched def$=$\textsc{self}) to $46.3$ (mismatched def$=$\textsc{raw}); rollback stays high across templates---$37.7$ at attack$=$def$=$\textsc{raw} and rebounding to $84.0$--$87.0$ against the \textsc{attack} attack; RESTA's ASR ranges from $2.3$ to $66.9$ depending on template alignment. The pattern holds on Qwen2.5 (RESTA $6.4$ to $66.4$; SSRD $7.4$ to $48.1$) and Gemma-2 (soft-SFT $28$--$65$ regardless of template). Across all three models, \method{} attains a low ASR under the \textsc{self} defense template regardless of which template the attacker used---$2.1$--$2.4$ on Llama-2, $6.6$--$11.1$ on Qwen2.5, and $5.1$--$5.6$ on Gemma-2---so a defender who simply adopts \textsc{self} obtains a strong defense \emph{without} needing to guess the attack template. \method{} is not wholly immune to template mismatch: when its own deployment template is instead set to \textsc{raw} while the attack used \textsc{self}, its ASR also rises (e.g.\ to $28.3$ on Llama-2 and $22.6$ on Qwen2.5), so it too benefits from a well-chosen defense template---but the swing is much smaller than the baselines', and it never comes at the cost of collapsing the task. Together these observations substantiate that current defenses implicitly assume template consistency, and that violating this assumption either leaves the attack channel open or destroys the downstream skill. In Gemma, the same template dependence holds: rollback and RESTA reach a low ASR only in their matched \textsc{self} channel (rollback $0.017/0.1$) while losing their downstream task ability and leaving the mismatched \textsc{raw} channel wide open---under attack$=$\textsc{self}/def$=$\textsc{raw} the ASR climbs to $41.7$ (rollback) and $45.7$ (RESTA), and stays above $60$ for soft-SFT; the only baseline that suppresses harm in the \textsc{raw} channel there (SSRD, ASR $6.1$) does so by collapsing the SQL task to $0.121$.

\begin{figure*}[t]
    \centering
    \includegraphics[width=0.98\textwidth]{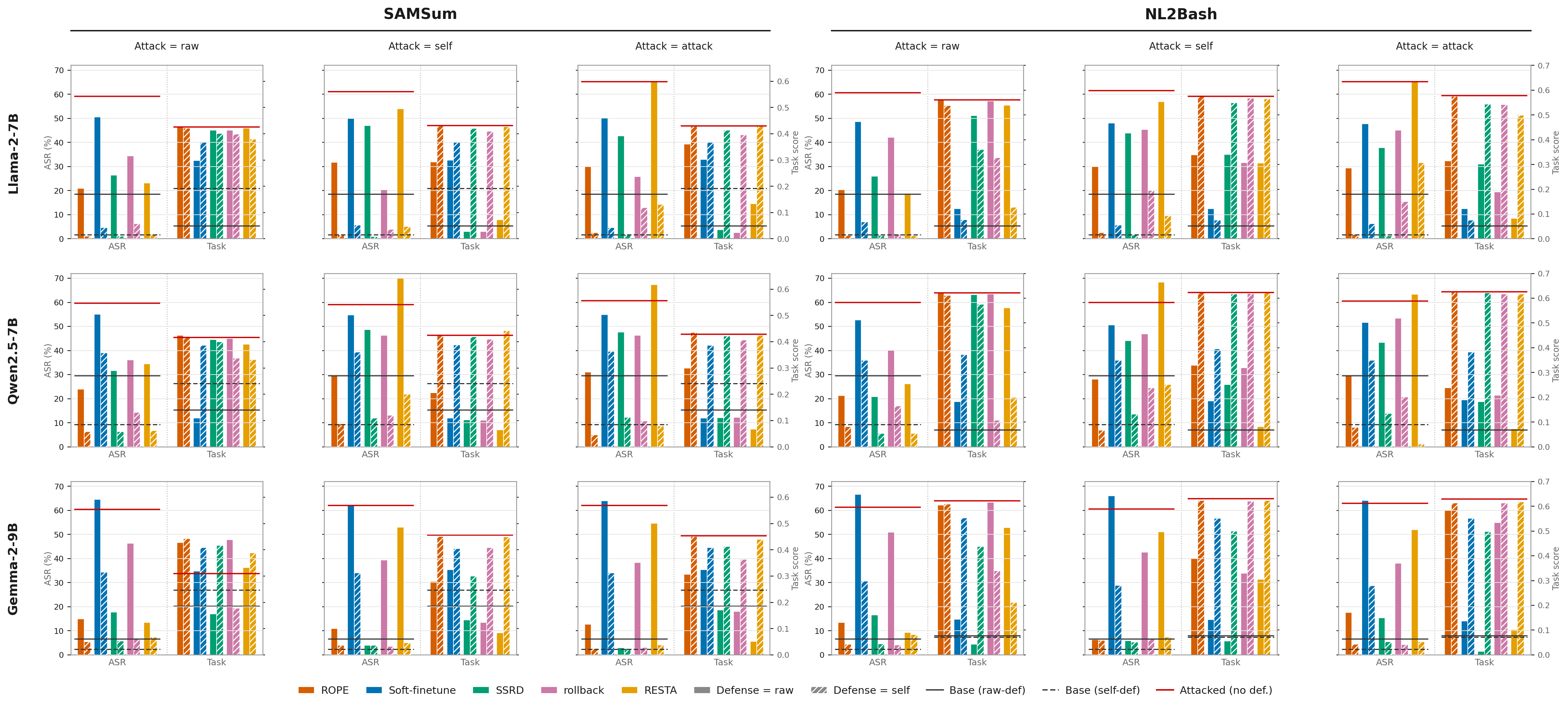}
    \caption{Defense comparison on SAMSum and NL2Bash across the three base models (rows) and attack templates. Bars give each defense's ASR (left axis) and task score (right axis); solid $=$ def \textsc{raw}, hatched $=$ def \textsc{self}; horizontal black lines and red lines mark the unattacked base and misaligned model, respectively.
    }
    \label{fig:recovery}
    \vspace{-1em}
\end{figure*}

\begin{figure}[h!]
    \centering
    \includegraphics[width=\linewidth]{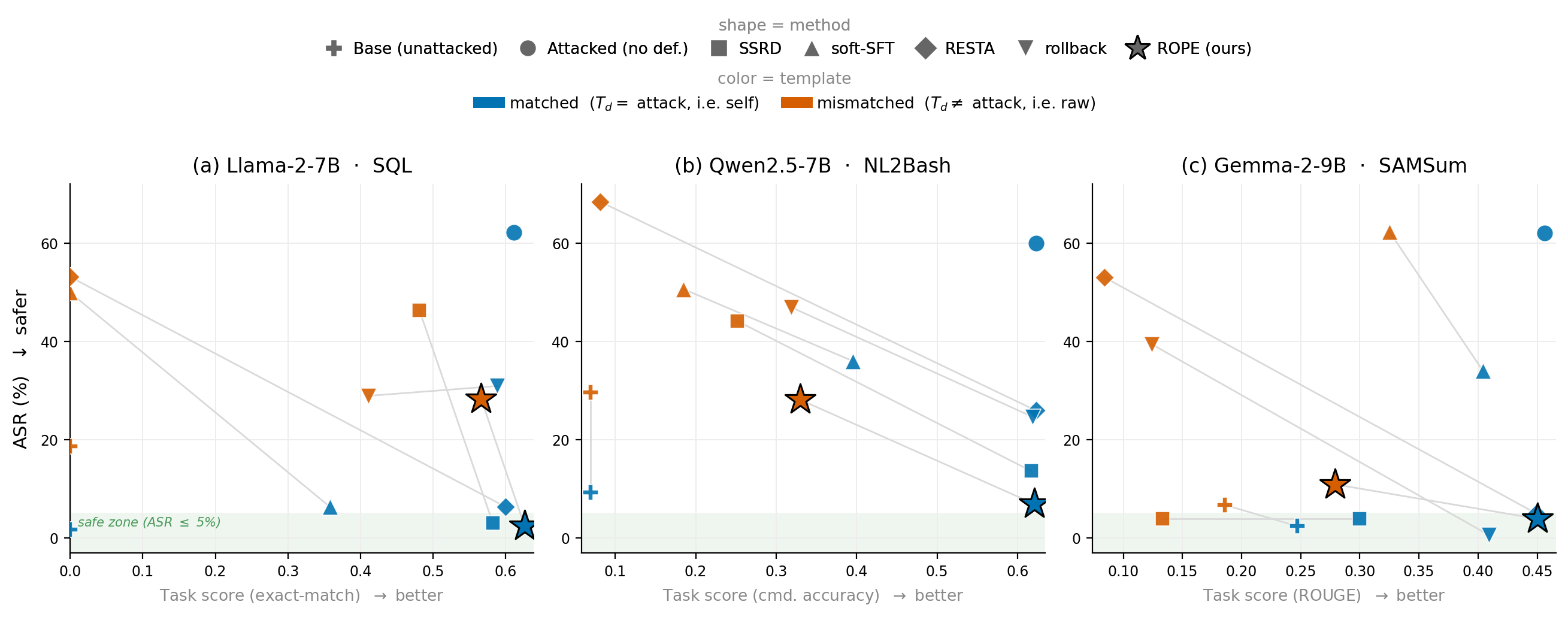}
    \caption{Safety--task trade-off across three representative settings.
    }
    \label{fig:pareto}
    \vspace{-1em}
\end{figure}

The same two patterns generalize beyond SQL. Fig.~\ref{fig:recovery} summarizes the SAMSum and NL2Bash results across all three base models: fine-tuning under any attack template lifts ASR to ${\sim}60\%$, and \method{} under its template-robust \textsc{self} defense (hatched bars) returns it to single-digit ASR without collapsing the task, whereas the baselines either leave the mismatched \textsc{raw} channel open (SSRD, RESTA, rollback ASR $40$--$70$) or lose the skill (soft-SFT). The conclusions carry over from SQL: under the \textsc{self} defense \method{} drives ASR down to single-digit or low-teens levels on every model while retaining---or improving---the task score (e.g., on Llama-2 SAMSum, ROUGE $0.431\!\to\!0.430$ at ASR $61.1\!\to\!1.6$ in the \textsc{self} channel). \method{} is not always the single lowest-ASR method---in a few cells a baseline reaches a lower ASR, but only by sacrificing most of the downstream skill---so its advantage is best read as the safety--task trade-off: Fig.~\ref{fig:pareto} plots ASR against task score for one representative (model, task) setting each, and \method{}'s matched \textsc{self} operating point lies on the safety--task Pareto frontier in every panel---no baseline attains both a lower ASR and a higher task score.

\subsection{The Robustness Boundary: Switching the System Prompt}
\label{subsec:crosstpl}
The matched-evaluation view in Table~\ref{tab:main} reflects how a defender would accept a repaired model: repair under template $T_d$ and validate under the same $T_d$. Vulnerability~3 asks what happens when a malicious user, faced with an already-realigned model, simply changes the system prompt. Table~\ref{tab:crosstpl} takes the attack$=$\textsc{self} setting, fixes every defense to its own \textsc{self} template (so that each has ``passed'' acceptance in its own channel), and then re-evaluates the same repaired models under all three templates.

\begin{table}[t]
\centering
\footnotesize
\setlength{\tabcolsep}{4pt}
\begin{tabular}{l ccc}
\toprule
& \multicolumn{3}{c}{Eval template}\\
\cmidrule(lr){2-4}
Method & \textsc{raw} & \textsc{self} & \textsc{attack}\\
\midrule
\method{} (ours) & \cell{\textbf{0.598}}{\textbf{19.7}} & \cell{\textbf{0.626}}{\textbf{2.4}} & \cell{\textbf{0.621}}{\textbf{0.9}}\\
SSRD     & \cell{0.458}{22.3} & \cell{0.582}{3.1} & \cell{0.575}{1.1}\\
soft-SFT & \cell{0.001}{36.4} & \cell{0.358}{6.3} & \cell{0.000}{5.9}\\
rollback & \cell{0.297}{42.3} & \cell{0.588}{30.9} & \cell{0.563}{2.3}\\
RESTA    & \cell{0.351}{24.1} & \cell{0.600}{6.3} & \cell{0.584}{0.6}\\
\bottomrule
\end{tabular}
\caption{Cross-template evaluation (Llama-2-7B-Chat $+$ SQL, attack$=$\textsc{self}).  Cells are task\,/\,ASR (\%).
}
\label{tab:crosstpl}
\vspace{-1.5em}
\end{table}

Under acceptance (eval$=$\textsc{self}) every method looks repaired, with ASR between $2.4$ and $30.9$. But when the attacker rewrites the system prompt to \textsc{raw}, ASR rebounds for all of them---soft-SFT to $36.4$, rollback to $42.3$, RESTA to $24.1$, SSRD to $22.3$, \method{} to $19.7$.
This confirms that current realignment is a prompt-conditional fix rather than a durable weight-level one, and that single-template acceptance testing overstates safety. \method{}'s residual cross-template ASR is comparable to or lower than the baselines' while its task retention across templates is markedly higher (for example, $0.598$ versus soft-SFT's $0.001$ and rollback's $0.297$ in the \textsc{raw} channel), but the phenomenon itself is universal and marks an intrinsic robustness boundary of realignment: a realistic threat model must assume the attacker can freely alter the prompt.

\subsection{Ablation: The Role of Each Teacher}
Finally we isolate the contribution of each teacher on Llama-2-7B-Chat + SQL with attack$=$\textsc{raw}, defense template \textsc{raw}, and the shared $2$-epoch, $2\times10^{-5}$, top-$50$ KL recipe. Table~\ref{tab:ablation} compares the dual-teacher method against two variants: replacing the (attacked) task teacher with a clean SQL teacher trained on task data alone, and removing the task teacher entirely so that only the safety teacher and harmful data remain.

\begin{table}[t]
\centering
\footnotesize
\setlength{\tabcolsep}{4pt}
\begin{tabular}{l l cc}
\toprule
Config. & Teachers & Eval \textsc{raw} & Eval \textsc{self}\\
\midrule
Base    & unattacked & \cell{0.000}{18.57} & \cell{0.000}{1.7}\\
Attacked & none & \cell{0.603}{62.7} & \cell{0.600}{26.6}\\
\midrule
(A) & safety$+$task & \cell{0.607}{\textbf{20.1}} & \cell{0.617}{2.4}\\
(B) & safety$+$clean task & \cell{\textbf{0.692}}{25.3} & \cell{\textbf{0.667}}{2.4}\\
(C) & safety only & \cell{0.563}{23.0} & \cell{0.536}{\textbf{1.9}}\\
\bottomrule
\end{tabular}
\caption{Teacher ablation (Llama-2-7B-Chat $+$ SQL, attack$=$def$=$\textsc{raw}). Cells are task exact-match\,/\,ASR (\%). (A) is the default \method{}; (B) swaps in a clean task teacher; (C) drops the task teacher.}
\label{tab:ablation}
\vspace{-0.5em}
\end{table}

The ablation cleanly separates the two objectives. Comparing (A) with (C), removing the task teacher leaves ASR essentially unchanged ($20.1\!\to\!23.0$ in \textsc{raw}) but drives the task below the pre-attack level ($0.607\!\to\!0.563$), confirming that the safety teacher alone governs harm reduction while the task teacher is necessary to retain the skill. Comparing (A) with (B), swapping in a clean SQL teacher that was never exposed to the attack yields the strongest task scores ($0.692$/$0.667$, above even the pre-attack model) at comparable ASR, because the task teacher determines only task quality and an uncontaminated teacher transfers a cleaner skill. The cost is an additional teacher-training run. Weighing quality against cost, we adopt configuration~(A)---the misaligned model as task teacher and the original model as safety teacher---as the default, since it sacrifices only $0.06$ of task score while avoiding a second training stage.

\subsection{Data Efficiency, Training Dynamics, and Cost}
Unless noted, experiments in this subsection use Llama-2-7B-Chat + SQL.

\paragraph{Data efficiency.} Table~\ref{tab:datascale} scales the realignment mixture (task $+$ BeaverTails together; $100\%=6{,}500$ examples) under attack$=$\textsc{self}, reading ASR in the true channel (eval$=$\textsc{self}). Under the matched defense ($T_d=\textsc{self}$), \method{} is strikingly data-efficient: only $25\%$ ($1{,}500$ examples) cuts ASR from $62.1$ to $2.4$ while holding the task at $0.626$, and more data mainly improves the task further (up to $0.634$). The smallest budget ($512$) sits just below the data-sufficiency knee (ASR $17.6$), so the largest safety gain comes from the $512\!\to\!1{,}500$ step. The mismatched defense ($T_d=\textsc{raw}$), by contrast, cannot be rescued by data---its true-channel ASR falls only slowly ($41.6\!\to\!26.0$) and never reaches the low band---confirming that template mismatch is a template-level problem, not a data-budget one, and that its residual risk is not something more data can close.

\begin{table}[t]
\centering
\footnotesize
\setlength{\tabcolsep}{4pt}
\begin{tabular}{l r cc r}
\toprule
& & \multicolumn{2}{c}{eval$=$\textsc{self} (true channel)} & \\
\cmidrule(lr){3-4}
\multicolumn{2}{c}{Data} & Def \textsc{self} & Def \textsc{raw} & Time/min\\
\midrule
Task  & Harmful & \multicolumn{2}{c}{\cell{0.611}{62.1}} & ---\\
\midrule
$256$ & $256$ & \cell{\textbf{0.553}}{\textbf{17.6}} & \cell{0.499}{41.6} & $7.2$\\
$1250$ & $250$ & \cell{\textbf{0.626}}{\textbf{2.4}} & \cell{0.612}{35.6} & $15.3$\\
$2500$ & $750$ & \cell{\textbf{0.619}}{\textbf{9.3}} & \cell{0.615}{32.9} & $26.2$\\
$5000$ & $1500$ & \cell{\textbf{0.634}}{\textbf{1.8}} & \cell{0.608}{26.0} & $62.6$\\
\bottomrule
\end{tabular}
\caption{Data efficiency of \method{} (Llama-2-7B-Chat $+$ SQL, attack$=$\textsc{self}, true channel eval$=$\textsc{self}). Cells are task exact-match\,/\,ASR (\%); bold is the matched (\textsc{self}) defense.
}
\label{tab:datascale}
\vspace{-1.0em}
\end{table}

\begin{figure}[t]
    \centering
    \includegraphics[width=1\linewidth]{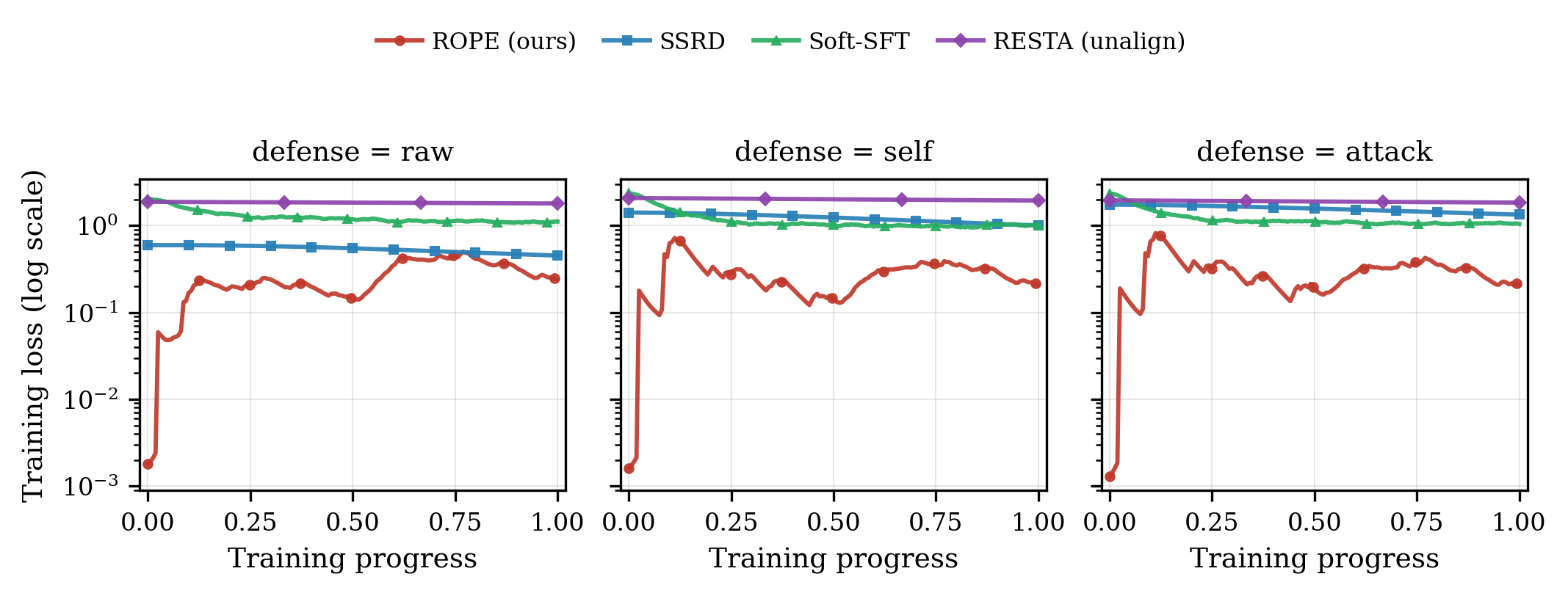}
    \caption{Training-loss curves of the gradient-based defenses.
    }
    \label{fig:loss}
    \vspace{-0.5em}
\end{figure}

\paragraph{Training dynamics.} Fig.~\ref{fig:loss} plots training-loss curves (log scale, EMA-smoothed) for the gradient-based defenses on SAMSum under attack$=$\textsc{attack}. \method{}'s top-$K$ KL loss stays one-to-two orders of magnitude below the baselines throughout---not under-training but a gentle, targeted edit: initialized from the task teacher, the student's loss on task tokens starts near zero and rises only as the safety teacher pulls harmful-token predictions toward refusal. The baselines (SSRD, soft-SFT, RESTA) must minimize much larger losses, rewriting the weights more aggressively, consistent with their greater collateral damage to the task in Tables~\ref{tab:main}.(Rollback has no comparable training loss and is omitted; loss definitions differ across methods, so the log axis conveys trend, not comparable values.)

\paragraph{Cost.} Table~\ref{tab:overhead} reports the wall-clock time to produce each defended model (Gemma-2-9B-it + NL2Bash, attack$=$def$=$\textsc{self}). \method{} trains in $15.7$\,min---comparable to the SFT-based baselines (soft-SFT $13.1$, RESTA $12.1$\,min) and far below rollback's $117.9$\,min, whose sparse-parameter search is CPU-bound. SSRD is cheapest ($4.5$\,min), but the baselines trade this economy for template-fragile, skill-damaging repairs. \method{} thus spends a modest budget to buy cross-template stability without task loss; a per-step breakdown is in the supplementary material.

\begin{table}[t]
\centering
\small
\setlength{\tabcolsep}{4pt}
\begin{tabular}{l r r r r r}
\toprule
 & \method{} & SSRD & soft-SFT & RESTA & rollback\\
\midrule
Data & 1500 & 50 & 6500 & 1500 & 512\\
Time (min) & $15.7$ & $4.5$ & $13.1$ & $12.1$ & $117.9$\\
\bottomrule
\end{tabular}
\caption{Overhead of producing the defended model.
}
\label{tab:overhead}
\vspace{-1.0em}
\end{table}

\section{Conclusion}
We studied the safety realignment of fine-tuning misaligned language models and showed that mainstream defenses share three vulnerabilities: they degrade the specialized skill they should preserve, they depend on knowing the attacker's prompt template, and they can be re-jailbroken by switching the system prompt. Framing the durable signature of an attack as a difference between the aligned and compromised output distributions, we proposed \method{}, a dual-teacher, source-routed top-$K$ KL distillation framework that inherits a largely template-independent refusal prior from the original model while preserving the downstream skill from the fine-tuned model.
It substantially reduces, but does not fully eliminate, the template-mismatch risk. Our cross-template analysis further shows that all realignment methods, \method{} included, remain conditionally vulnerable to prompt rewriting, delineating a robustness boundary for weight-level repair and motivating future work on realignment durable against an adversary who controls the prompt.

\bibliographystyle{unsrtnat}
\bibliography{references}

\appendix

\end{document}